\documentclass{article}

\usepackage[preprint]{arxiv}
\pdfoutput=1

\usepackage[utf8]{inputenc} 
\usepackage[T1]{fontenc}    
\usepackage{hyperref}       
\usepackage{url}            
\usepackage{booktabs}       
\usepackage{amsfonts}       
\usepackage{nicefrac}       
\usepackage{microtype}      
\usepackage{graphicx}
\usepackage{amsmath}
\usepackage{multirow}
\usepackage{algorithm}
\usepackage{algorithmic}
\usepackage{subfigure}
\usepackage{caption}
\usepackage{wrapfig}
\usepackage{cleveref}

\title{Label Smoothing and Adversarial Robustness}

\author{%
  Chaohao Fu\\
  Dept. of Computer Science and Engineering\\
  Shanghai Jiao Tong University\\
  \texttt{chhfu-1996@sjtu.edu.cn} \\
   \AND
   Hongbin Chen\\
   Dept. of Computer Science and Engineering\\
   Shanghai Jiao Tong University\\
   \texttt{k160438@sjtu.edu.cn} \\
   \And
   Na Ruan \thanks{Corresponding author}\\
   Dept. of Computer Science and Engineering\\
   Shanghai Jiao Tong University \\
   \texttt{naruan@cs.sjtu.edu.cn} \\
   \And
   Weijia Jia \footnotemark[1]\\
   BNU-UIC Institute of Artificial Intelligence and Future Networks\\
   Beijing Normal University (BNU Zhuhai)\\
   BNU-HKBU United International College\\
   Shanghai Jiao Tong University\\
   \texttt{jia-wj@cs.sjtu.edu.cn} \\
}

\begin{document}

\maketitle

\begin{abstract}
Recent studies indicate that current adversarial attack methods are flawed and easy to fail when encountering some deliberately designed defense. Sometimes even a slight modification in the model details will invalidate the attack. We find that training model with label smoothing can easily achieve striking accuracy under most gradient-based attacks. For instance, the robust accuracy of a WideResNet model trained with label smoothing on CIFAR-10 achieves 75\% at most under PGD attack. To understand the reason underlying the subtle robustness, we investigate the relationship between label smoothing and adversarial robustness. Through theoretical analysis about the characteristics of the network trained with label smoothing and experiment verification of its performance under various attacks. We demonstrate that the robustness produced by label smoothing is incomplete based on the fact that its defense effect is volatile, and it cannot defend attacks transferred from a naturally trained model. Our study enlightens the research community to rethink how to evaluate the model's robustness appropriately.
\end{abstract}

\section{Introduction}
Deep neural networks have become increasingly effective at various tasks. When training a network, label smoothing is a commonly used "trick" to improve deep neural network performance, which uses smoothed uniform label vectors in place of one-hot label vectors when computing the cross-entropy loss during training. Szegedy et al.\cite{szegedy2016rethinking} originally proposed label smoothing as a regularization strategy that improved the performance of
the Inception architecture on the ImageNet dataset. Since then, a set of works\cite{zoph2018learning,real2019regularized,huang2019gpipe} on image classification task start to use this means of regularization to improve generalization. They think label smoothing is a form of output distribution regularization that prevents overfitting of a neural network by softening the ground-truth labels to penalize overconfident outputs\cite{pereyra2017regularizing}. This method has been extended to many fields, including speech recognition \cite{chorowski2017towards}, machine translation\cite{vaswani2017attention}. The recent work have also revealed that label smoothing improves model calibration\cite{muller2019does}.

Over the last few years, there has been increasing works on the neural network's brittleness. Szegedy et al.\cite{szegedy2013intriguing} first revealed that the existence of adversarial examples, concretely, imperceptible changes to an image can cause a DNN to mislabel it. Since then, researchers in the field have launched arm races against attack and defense. Among attack methods, a series of methods are commonly used, which exploits gradient information of models to construct or search for the direction of perturbation. We call them gradient-based methods, including PGD\cite{madry2017towards}, CW\cite{carlini2017towards} and so on. 

However, recent studies have shown that evaluating robustness by these attack methods is insufficient and lead to overestimation of robustness \cite{mosbach2018logit,croce2019scaling}. Inspired by them, we find that training model with label smoothing can invalidate most gradient-based attacks. The robust accuracy of the label smoothing model exceeds most commonly accepted defense models. We investigate robust and natural accuracy of models on various datasets with and without label smoothing. From Table \ref{tab1}, we can clearly see that model using label smoothing has a significant defensive effect against adversarial attacks no matter on what dataset. 

\begin{table*}[ht]
\caption{Natural accuracy and robust accuracy of differernt models trained with and without label smoothing.}
\label{tab1}
\centering
\begin{tabular}{ccccc}
    \toprule
    Data set & Architecture & Circumstance & Accuracy(with LS) & Accuracy(without LS) \\
    \midrule
    \multirow{2}{*}{MNIST} & \multirow{2}{*}{4-layer CNN} & natural & 99.11 & 99.73 \\
    &&FGSM & 99.03 & 0\\
    \midrule
    \multirow{2}{*}{CIFAR-10} & \multirow{2}{*}{WideResNet-34-10} & natural & 94.74 & 95.44 \\
    &&PGD &75.10&0\\
    \midrule
    \multirow{2}{*}{CIFAR-100} & \multirow{2}{*}{WideResNet-34-10} & natural & 80.50 & 81.10\\
     &&PGD &34.23&0.03\\
    \bottomrule
\end{tabular}
\end{table*}


Note that even for the adversarial training model \cite{madry2017towards} computational cost is required to achieve adversarial robustness. Previous work proves that training adversarially robust model is particularly difficult\cite{madry2017towards}. Either the network requires more capacity to be robust\cite{madry2017towards,nakkiran2019adversarial}, or the models tend to require more data\cite{alayrac2019labels,carmon2019unlabeled,najafi2019robustness}. Therefore, neither change the network capacity nor use more data, only to modify the training labels makes the model have a defensive effect against adversarial attacks, which make us raise following question:

\begin{quote}
\centering
    \textit{Does label smoothing bring real robustness to the DNNs?}
\end{quote}

In order to explore this problem, we study the relationship between label smoothing and adversarial robustness from both theoretical and experimental aspects. Our contributions are as follows:
\begin{itemize}
    \item We indicate that label smoothing has a very significant effect to most gradient based attack.
    \item We analyze the reasons why the label smoothing model invalidates most gradient-based attacks theoretically.
    \item We demonstrate that label smoothing can not bring stable robustness. We study the properties of the label smoothing model and summarize the key points to break through its robustness experimentally.
\end{itemize}

Note that attacks that can be used to assess robustness should either be able to systematically find adversarial examples or be able to declare with high confidence that no adversarial examples exists. Our work reveal the necessity and urgency of proposing  systematic and efficient attack methods to evaluate the robustness of neural networks.

\section{Theoretical Analysis}

In this section, we will analyze how label smoothing training change the model. We will first make a theoretical analysis, and then give a toy example of why such changes will affect the robustness of the model.

\subsection{Preliminaries}
A N-layer DNN is a mapping $f: \mathbb{R}^{n} \mapsto \mathbb{R}^{m}$ that accepts an input $\boldsymbol{x} \in \mathbb{R}^{n}$ and produces an output $f(\boldsymbol{x}) \in \mathbb{R}^m$ with parameter $\theta$.  It can be expressed as: 
\begin{equation}
    f(\boldsymbol{x}) = F^{(N)}\left(F^{(N-1) \ldots}\left(F^{(1)}(\boldsymbol{x})\right)\right) \label{eq1}
\end{equation}
where $F^{(i+1)}(x)=\sigma\left(w_{i} \cdot F^{(i)}(x)+b_{i}\right)$ for non-linear activation function $\sigma (\cdot)$, model weights $w_i$ and bias $b_i$. 
Let $Z$ denote the outputs of $f$, $S$ denote the output of the softmax layer. that is, $Z=f(\boldsymbol{x})$ and $S = \operatorname{softmax}(Z)$. Let $z_i$ denote the logit of $i$-th class, and $s_i$ is the $i$-element of $S$, which satisfies $0 \leq s_i \leq 1$ and $s_1 + s_2 + ... + s_m = 1$ due to the property of softmax. The DNN assigns the label $\hat{y}=\arg \max _{i} f(\boldsymbol{x})$.

For a network trained with hard labels for a classification task with $K$ classes, we use one-hot label vectors $\boldsymbol{y}$, where $y_k$ is "1" for the correct class and "0" for the rest. For a network trained with label smoothing, we use the soft label vectors $\boldsymbol{y}^{LS}$ instead. The value of each dimension is modified according to $y_{k}^{L S}=y_{k}(1-\alpha)+\alpha / K$. The parameter $\alpha$ which we call smoothing fator measures the extent of smoothness. When $\alpha$ is small, the gap between wrong and correct class is relatively lager, smoothness level is low. As the $\alpha$ increase, the smoothness of the label vector increases, and the gap between the correct and wrong class decreases.

\subsection{Impacts of Label Smoothing}
Most of the neural networks are trained with the Cross Entropy (CE) as the objective loss: $\mathcal{L}_{ce}=\sum_{i=1}^{K}-y_{k} \log \left(s_{k}\right)$, where $K$ is the number of classes. The loss can be written as follows when $y$ is one-hot label:
\begin{equation}
    \mathcal{L}_{ce}(y, s) = -\log(s_t) \label{eq2}
\end{equation}
where $s_t$ is output probability of correct class. And if we smooth the label according to $y_{k}^{L S}=y_{k}(1-\alpha)+\alpha / K$, the CE loss will become:
\begin{equation}
\begin{split}
    \mathcal{L}_{ce}(y^{LS}, s) &=\sum_{i=1}^{K}-y^{LS}_{k} \log \left(s_{k}\right) \\
&=-(1-\alpha+\alpha / K) \log s_{t}-\alpha / K \sum_{i \neq t} \log s_{i}
\end{split}\label{eq3}
\end{equation}

By substituting $s$ in Equation \eqref{eq2} and \eqref{eq3} using $s_{i}=e^{z_{i}} / \sum_{j=1}^{K} e^{z_{j}}$, we can further deduce the relationship between loss function and logits when using one hot label and smoothed label respectively. 

\begin{align}
    &\mathcal{L}_{ce}(y, z) = \log \left(1+\sum_{i \neq t} e^{z_{i}-z_{t}}\right) \label{eq4}\\
    &\mathcal{L}_{ce}(y^{LS}, z) = \frac{\alpha}{K} \sum_{i \neq t}\left(z_{t}-z_{i}\right) +\log \left(1+\sum_{i \neq t} e^{z_{i}-z_{t}}\right) \label{eq5}
\end{align}


Let $M_i = z_{t}-z_{i}$ denote the margin between the corresponding logit of the correct class and one of the other classes, then,
\begin{align}
    &\mathcal{L}_{ce}(y, z) = \log \left(1+\sum_{i \neq t} e^{-M_i}\right) \label{eq6}\\
    &\mathcal{L}_{ce}(y^{LS}, z) = \frac{\alpha}{K} \sum_{i \neq t}M_i +\log \left(1+\sum_{i \neq t} e^{-M_i}\right) \label{eq7}
\end{align}
Traditional networks trained with hard labels try to decrease the Equation \eqref{eq6} to increase the margin of the correct logit and other logits. Note that there is one more item on the right side of Equation \eqref{eq7} than Equation \eqref{eq6}, when label smoothing is applied to the label, a term opposite to the original optimization objective (reducing $M$) is added to the loss function. Thus, the original loss will encourage the margin to grow infinitely, the intensity of encouragement (which is gradient) will gradually decrease and eventually converge.  By contrast, when the label smoothing is applied, the first term of Equation 6 will suppress the growth of margin. The ultimate effect of the two losses restraining each other is to reach a compromise in a certain range acceptable to both terms. In other words, the value range of $M_i$ is limited to an acceptable range of two terms of loss. Calculate the partial derivative of loss with respect to a specific $M_u$:
\begin{equation}
    \frac{\partial \mathcal{L}_{ce}}{\partial M_{u}}=\frac{\alpha}{K}-\frac{e^{-M_{u}}}{1+\sum_{i} e^{-M_{i}}} \label{eq8}
\end{equation}
According to the theorem that the partial derivative at the extreme point is 0, we can obtain the approximate range of $M$. More ideally, the value of $M$ is a constant and satisfies $M_1 = M_2 =...M_k=m$, and depending on the properties of the softmax function, we can conclude that logits also have a range, and the size of this range is determined by $\alpha$. The approximate range of logits are listed in Table \ref{tab2} according to the experiment.

\begin{table}[ht]
    \caption{The approximate ranges of logit for models with different smoothing factor $\alpha$.}
    \label{tab2}
    \centering
    \begin{tabular}{c|c||c|c}
    \toprule
    $\alpha$&range&$\alpha$&range\\
    \midrule
        0 &  [-5.00, 15] & 0.5 & [-0.30, 2.2]\\
        0.1 & [-0.55, 4.5] & 0.6 & [-0.25, 2.0]\\
        0.2 & [-0.45, 3.5] & 0.7 & [-0.20, 1.6]\\
        0.3 & [-0.40, 3.0] & 0.8 & [-0.14, 1.2]\\ 
        0.4 & [-0.35, 2.5]& 0.9 & [-0.08, 0.7]\\
        \bottomrule
    \end{tabular}
\end{table}

\subsection{Toy Example}
Based on the results of the above theoretical analysis and experimental validation, we try to use a simplified model to explain why label smoothing has an impact on the effectiveness against adversarial attacks.

There are two loss functions commonly used in the algorithm of adversarial example generation. One is the cross-entropy loss, the other is margin loss:

\begin{align}
&\hat{\mathcal{L}}_{ce}(y, z) = \log \left(1+\sum_{i \neq t} e^{z_{i}-z_{t}}\right)  &\text { cross-entropy loss } \label{eq9}\\
&\hat{\mathcal{L}}_{mg}(y, z)=-z_{t}+\max _{i \neq t} z_{i} &\text { margin loss }\label{eq10}
\end{align}

Regardless of the loss function, the objective is to use the inverse direction of the gradient as the direction of the search for the adversarial example. Normally along this direction, the logit of the correct class decreases, and the predicted label changes when the search effort crosses the decision boundary ($z_t= z_i$). If the perturbation size meets the requirements (within $\epsilon$-radius $L_p$ ball) at this point, the adversarial example is successfully found.

\begin{figure}
    \centering
    \includegraphics[width=0.45\columnwidth]{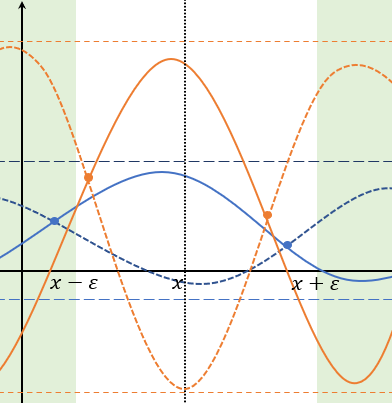}
    \caption{Illustration of how label smoothing impact attack. Due to the decision boundary points of blue model are farther apart, the number of boundary points within the permissible range of perturbation is reduced, resulting in the inability to find adverarial example based on gradients sometimes.}
    \label{fig1}
\end{figure}

Let us take a toy example of what will happen when label smoothing is applied.  Consider a one-dimension classification task, the relationship between the logit of the network output and the input is shown in Figure \ref{fig1}. The solid line represents the correct logit, and the dotted line represents one of the remaining logit. The orange line represents the model with no label smoothing applied, and the blue line represents the applied. Empirically, the two models achieve extremes at similar locations, and the range of logit is smaller due to label smoothing applied, resulting in the different location of the decision boundary points between two models. 
Two different decision points of the model are far apart, which leads to fewer decision boundary points within the allowable perturbation interval. As the Figure \ref{fig1} shows, Assuming that we use the logit of the correct class as a loss function to generate adversarial examples, we look for adversarial examples in the direction of the descent of the correct class. When $x' \in [x-\epsilon, x]$, this method fails to generate adversarial example.

\section{Experiment Verification}
It has been revealed that most of attack methods for evaluation of defense model are often insufficient or give a wrong impression for robustness over the past years. 
We have demonstrated that label smoothing greatly improves the model's performance against the PGD attack. However, it is necessary to have a more comprehensive understanding of the robustness of the label smoothing model. In this section,  we use more representative attack methods to test the robustness more comprehensively and try to explain the experiment results with the conclusion of the previous section. 

\subsection{Experimental Setup}

We consider the classification task on CIFAR-10, and all the model's architecture is WideResNet-34-10. For $L_\infty$ bounded attak, we use maximum perturbation $\epsilon = 8/255$; for $L_2$ bounded attack, $\epsilon=140/255$. We investigate two models that are the most representative defensive models as the baseline. Standard adversarial training (Madry et al.\cite{madry2017towards}) and TRADES (Zhang et al.\cite{zhang2019theoretically}). Both of them are typical models based on adversarial training. The specific training settings for both models are the same as their original paper.


\begin{wrapfigure}{r}{7cm}
\centering
\includegraphics[width=0.5\columnwidth]{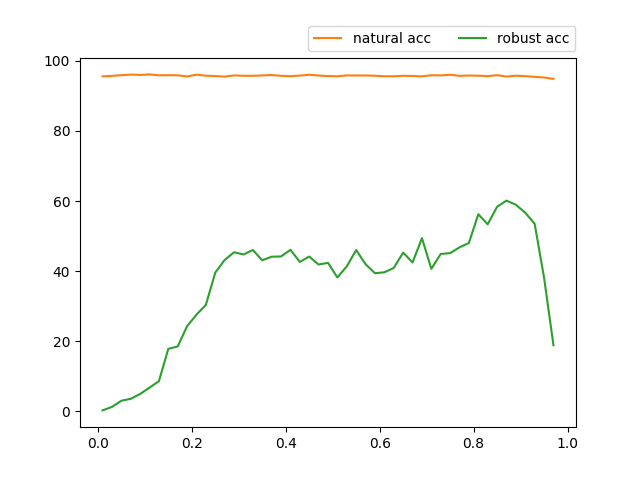} 
\caption{Accuracy of model trained with Label smoothing for different smoothing factors.}
\label{fig2}
\vspace{-20pt}
\end{wrapfigure}

To avoid much concentration on optimizing the hyper-parameter $\alpha$ of label smoothing model, we fix it as a constant. Therefore, We first test when the label smoothing model has the best defense performance. We train a series of networks with different smoothing factors and evaluate them under the same attack setting: maximum perturbation $\epsilon= 8/255$, step size $\alpha = \epsilon/10$, and total perturbation steps $K=20$. The results are shown in Figure \ref{fig2}.

From Figure \ref{fig2}, we can intuitively observe that when the smoothing factor is greater than 0.3, the model starts to have a remarkable defense effect against PGD attacks and reaches the maximum value when the factor is 0.9, then decreases.
Based on the analysis in the previous section, this pattern of change is logical. Thus, in all the next experiments, the smoothing factor of all label smoothing models is set to 0.9 as it reaches the best performance.

We consider evaluation models in 4 aspects. These are PGD-like attack, CW attack, our proposed attack and transferability test. Details of the experimental settings for each section will be described later.

\subsection{PGD Attack and Its Variants}
Currently, PGD attack is a mostly used algorithm to test adversarial robustness. This method has the advantage of computationally cheap and performs well in many cases. However, it has been shown that PGD will lead to a significant overestimation of robustness. Recent years, a line of works about PGD variants are proposed. We first give a definition of PGD-like attacks:

Given a classifier $f$, original example $\boldsymbol{x}_{orig}$, original target $t$, adversarial loss function $\mathcal{L}_{f,t}(\boldsymbol{x})$. The PGD-like algorithms refer to the algorithms that generate adversarial example by performing the following procedure one or more iterations.  
\begin{itemize}
    \item Starting disturb at point $x_0$ which satisfies \\ 
    \begin{equation}
        \boldsymbol{x}_0 \in \mathcal{B}_\epsilon^p(\boldsymbol{x}_{orig})\label{eq11}
    \end{equation}
    \item Searching new point by \\ 
    \begin{equation}
        \boldsymbol{x}_{k+1} = \Pi_{\mathcal{B}_\epsilon^p(\boldsymbol{x}_{orig})}(\boldsymbol{x}_k + \eta \text{sign}(\nabla \mathcal{L}_{f,t}(\boldsymbol{x}_{k})))\label{eq12}
    \end{equation}
\end{itemize}
where $ \mathcal{B}_\epsilon^p(\boldsymbol{x}_{orig}) $ represents $L_p$-ball centered on $\boldsymbol{x}_{orig}$, $\eta$ is the step size and $\Pi_{\mathcal{B}_\epsilon^p(\boldsymbol{x}_{orig})}$ indicates the operation of projection to the surface of $L_p$-ball.

We can summarize three characteristics of the PGD-like attack algorithm from the above definition.
\begin{itemize}
    \item The starting point of perturbation is in the neighborhood of a certain norm space of the original point.
    \item The direction of perturbation is the sign direction of the adversarial loss function.
    \item Use projection operations to control boundaries.
\end{itemize}

\begin{table*}[ht]
    \caption{The components and specific settings of different PGD-like attack algorithms.}
    \label{tab3}
    \centering
    \begin{tabular}{c|cccc}
         \toprule
         & Restarts & Initialization & Steps \& Step size & Loss function \\
         \midrule
         FGSM& w/o & original& 1, $2\epsilon$ & cross-entropy\\
         $\text{PGD}^{10}$& w/o & random & 10, $\epsilon/5$ & cross-entropy\\
         $\text{PGD}^{20}$& w/o & random & 20, $\epsilon/10$ & cross-entropy\\
         $\text{PGD}^{40}$& w/o & random & 40, $\epsilon/20$ & cross-entropy\\
         PGD-CW & w/o&random & 20, $\epsilon/10$ & margin\\
         MT& 18 & random & 20, $\epsilon/10$ & different margin\\
         ODI& 20 & calculated & 20, $\epsilon/10$ & margin\\
         AA& 1 & random & 100, auto & cross-entropy\\
         \bottomrule
    \end{tabular}
\end{table*}

\begin{table*}[ht]
    \caption{Robustness evaluation of different models by various PGD-like attack. We report clean test accuracy, the robust accuracy of the various PGD-like attack. Attack algorithms are divided into two categories: the classic PGD-like attack and the latest improved PGD-like attack}
    \label{tab4}
    \centering
    \begin{tabular}{c|c|ccccc||ccc}
        \toprule
        Defense & clean & FGSM & $\text{PGD}^{10}$ & $\text{PGD}^{20}$ & $\text{PGD}^{40}$ & PGD-CW & MT & ODI & AA\\
        \midrule
         Madry&86.83&56.88&52.32&51.24&51.14&50.73&50.34&49.13&49.35\\
         Zhang&84.92&64.87&56.12&55.09&55.89&53.69&52.55&52.71&53.15\\
         LS&94.74&67.17&64.01&74.83&73.15&74.10&25.02&0.15&6.86\\
         \bottomrule
    \end{tabular}
\end{table*}

We mainly apply two classic methods (FGSM and PGD) and three recent proposed methods(MT, ODI and AA)
to test the adversarial robustness of the label smoothing model. These three new attack algorithms are considered to produce more accurate robustness evaluations than traditional PGD. MT represents the MultiTargeted attack \cite{gowal2019alternative}, ODI represents Output Diversified Initialization attack \cite{tashiro2020ods}, AA represents AutoAttack \cite{croce2020reliable}.

We first explain how these attacks are distinguished from each other and following the same pattern mentioned above, and introduce their specific attack settings in following experiment. Firstly, FGSM and PGD both perform the above procedure once without restarts. FGSM is the simplest PGD-like algorithm because it only perturbs one step from starting point $\boldsymbol{x}_{orig}$, but the step size $2\epsilon$ is also the largest. For PGD attack, we will test three different attack settings with different steps $k$ and step size $2\epsilon/k$, which are denoted as $\text{PGD}^{k}$; PGD-CW attack is a way to realize CW attack in most defense papers. In this implementation, only the cross-entropy loss function in PGD attack is replaced by the margin loss function. These attacks' starting points are random sampled from $\epsilon-$radius $L_p$ ball $\mathcal{B}_\epsilon^p(\boldsymbol{x}_{orig})$. MT attack disturbs with multiple restarts and picks a new target class logit to calculate margin loss at each restart. ODI provides a more effective initialization strategy to determine the starting point with diversified logits at each restart. AA attack is a parameter-free ensemble of four attacks: FAB, two proposed Auto-PGD attacks with different loss functions, and the black-box Square Attack. But here, we also use one of its Auto-PGD attacks---APGD-CE, as a PGD variant. APGD-CE improves the attack effect through searching good initial point with restarts and a step size selection strategy. The components and specific settings of these attacks are listed in Table \ref{tab3}.

The results are shown in three parts as Table \ref{tab4}. Firstly, label smoothing model reaches the best natural accuracy since no noise is added to the training samples.
Secondly, under all the classic PGD-like attacks, including FGSM and PGD, the label smoothing model has also achieved the highest defense effect in comparison to two adversarial training models. It should be noted that the accuracy of the two models based on adversarial training does not fluctuate much when step increases (step size decrease). However, the label smoothing model presents an intuitive rising trend. Thirdly, 3 recent proposed attacks reduce the accuracy of the label smoothing model enormously, while the accuracy of two adversarial training models does not decrease much.

Synthesize the performance of the label smoothing model under differnt PGD-like attack, it can empirically conclude that label smoothing model does not have stable robustness. Firstly, The robustness of the label smoothing model is readily affected by the step size and step number. It can be an explanation that the larger step size is beneficial to jump out of the local wrong perturbation direction. Secondly, 3 new improved PGD-like attacks almost destroyed the defense of the model, though to varying degrees. It mainly owes to their restart mechanism. The restart mecanism it to try more perturbation directions and choose the best. MT utilize different classes of margin loss to vary the direction, ODI utilize different initial point to vary the direction and AA utilize the larger step size to vary the direction. Among them, the method of changing the starting point is the best, increasing step size is the second, and the effect of changing the loss function is the worst. In addition, the result of PGD-CW attack manifests that it is not critical whether the loss function is cross-entropy or margin.

\subsection{CW Attack} 
CW attack is proposed by Carlini et al.\cite{carlini2017towards}. As mentioned above, the PDG-CW attacks are not strictly implementation of CW attacks, according to the algorithm proposed by Carlini et al.\cite{carlini2017towards}. Find an adversarial example for image $x$ within $L_p$ distance $\delta$ with CW attack is equivalent to solving the following optimization problems,

\begin{equation}\label{eq13}
    \begin{aligned}
    &\text { minimize }\|\delta\|_{p}+c \cdot l(x+\delta) \\
    &\text { such that } x+\delta \in[0,1]^{n}
\end{aligned}
\end{equation}

\noindent where $c>0$ is hyper-parameter and the best $l$ is the margin loss according to the original paper. CW attacks differ from gradient-based PGD-like attacks in the following points. 
\begin{itemize}
    \item CW's starting point can be any point, not necessarily within $\epsilon-$radius $L_p$ ball .
    \item CW use variable substitution to control boundaries, instead of projection operations.
    \item CW's perturbation is gradient direction rather than gradient sign direction. 
    \item CW need parameter fine-tuning for better attack effect, including $c$ and step number and step size etc.
\end{itemize}

In order to reduce the number of influence factors, we only show the results under $L_2$ norm here and set up the hyper-parameter. The First row in Table \ref{tab5} reports that under the CW attack, label smoothing shows poor robustness, while the two adversarial training models still maintain certain robustness.

Note that in the first experiment, $\boldsymbol{x}_0$ was initialized to 0. To investigate the influence of the starting point on results, we initialize $\boldsymbol{x}_0$ to the original image and test again. The results show that the accuracy of the label smoothing model is improved after the starting point is changed.

\begin{table}[ht]
    \caption{Accuracy of different model under CW attack using original algorithm. Distance metric is $L_2$ norm, $\epsilon = 140/255$, $c=0.1$, confidence bias is -50. The starting point of the first row is the black image($\boldsymbol{x}_0=0$), and the starting point of the second row is the original image ($\boldsymbol{x}_0 =\boldsymbol{x}_{orig}$).}
    \label{tab5}
    \centering
    \begin{tabular}{c|ccc}
         \toprule
         Starting point& LabelSmoothing&Madry&Zhang \\
         \midrule
         $\boldsymbol{x}_0=0$& 8.49&42.12&51.29\\
         $\boldsymbol{x}_0 =\boldsymbol{x}_{orig}$&37.83&43.14&52.73\\
         \bottomrule
    \end{tabular}
\end{table}

From the above experiments on CW attack, It can conclude that label smoothing model can be broken by CW attack which is strictly executed according to the original paper, and the key point to break is the selection of the starting point. In view of the fact that the optimization target of CW attack introduces the term of perturbation size and the arbitrary selection of starting point, it is not surprising that the defense effect from label smoothing failure.

\subsection{Our Attack}
We propose a novel $L_\infty$ attack algorithm, which combines the advantages of both PGD and CW. This method uses the idea of variable substitution in CW attack and optimization objective in PGD attack. For $L_\infty$ norm bounded attack, The value of each pixel changes in [$-\epsilon$, $\epsilon$], while pixel value itself have range [0, 1]. Combine them together, each pixel value $x$ satisfies:
\begin{equation}
    x \in \{ x | \max \{x-\epsilon, 0\}\leq x \leq\min \{x+\epsilon, 1\} \}\label{eq14}
\end{equation}
We propose a new variable substitution to map this finite interval to an infinite interval. Substitute $\boldsymbol{x}$ by variable $\boldsymbol{a}$.
\begin{equation}
    \boldsymbol{a} = \left(\boldsymbol{x}_{max} - \boldsymbol{x}_{min}\right)*(1/2 (\tanh{\boldsymbol{w}} + 1)) + \boldsymbol{x}_{min}\label{eq15}
\end{equation}
Thus, we obtain an optimizable parameter $\boldsymbol{w}$ without range restriction and the generation of adversarial example turn into an unconstrained optimization problem: $\text { minimize } \mathcal{L}(\boldsymbol{a})$. We can solve this problem by gradient decent algorithms and then substituting the solution $\arg\min_{\boldsymbol{w}}\mathcal{L}(\boldsymbol{a})$ to Equation to gain the adversarial example $\boldsymbol{x}_{adv}$. A detailed description of our attack can be found in Algorithm \ref{algo1}.


\begin{algorithm}
    \caption{Our $L_\infty$ attack algorithm.}
    \label{algo1}
    {\bf Input:}
    The initial image, $\boldsymbol{x}$;label $y$, model $f$.\\
    {\bf Output:}
    Adversarial example $\boldsymbol{x}_{adv}$. \\
    {\bf Parameters:}
    Perturbation bound $\epsilon$, step size $\alpha$, number of steps $K$. \\
    \begin{algorithmic}[1]
        \STATE Initialize $\boldsymbol{x}_{adv} = \boldsymbol{x}$;
        \STATE Initialize $\boldsymbol{w}_{0} = 0$;
        \STATE Calculate $\boldsymbol{a}_0$ by Equation \eqref{eq15};
        \FOR {$k \in \{1,...,n\}$}
        \STATE $\boldsymbol{w}_{k+1} = \boldsymbol{w}_{k} - \eta \nabla_{\boldsymbol{w}}\mathcal{L}(\boldsymbol{a}_k)$
        \STATE Calculate $\boldsymbol{a}_{k+1}$ by Equation \eqref{eq15}.
        \IF{$\ell\left(\boldsymbol{x}_{adv}\right) < \ell\left(\boldsymbol{a}_{k+1}\right)$}
        \STATE $\boldsymbol{x}_{adv} \leftarrow \boldsymbol{a}_{k+1}$
        \ENDIF
        \ENDFOR
        \RETURN {$\boldsymbol{x}_{adv}$}
    \end{algorithmic}
\end{algorithm}

We use algorithm 1 to do ablation study on step numberand step size, since they are two independent factors in this algorithm. 

\begin{figure*}[ht]
    \centering
    \subfigure{
        \centering
        \includegraphics[width=0.48\columnwidth]{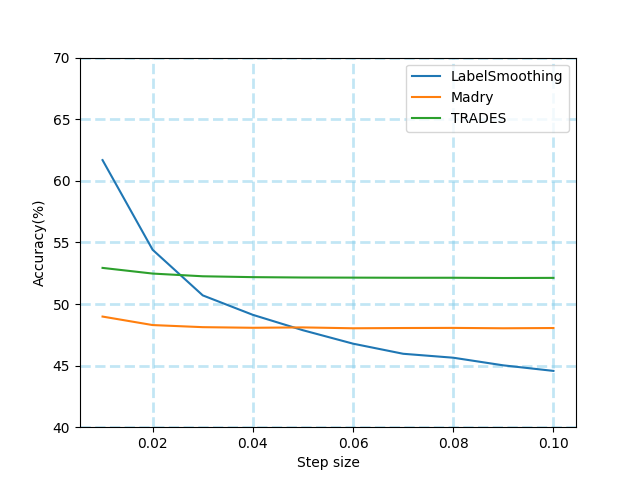}
    }
    \subfigure{
        \centering
        \includegraphics[width=0.48\columnwidth]{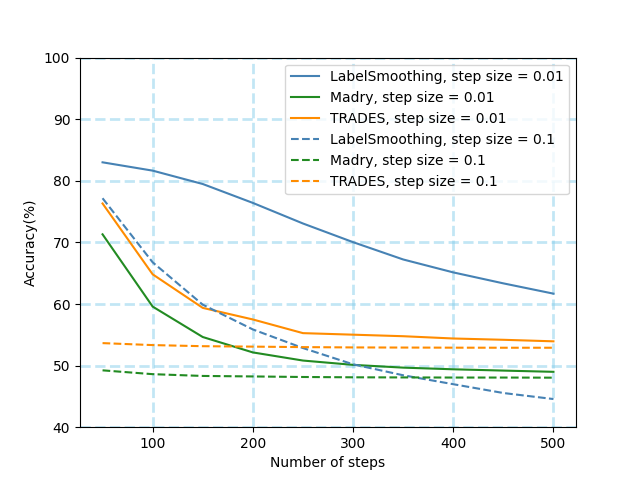}
    }
    \caption{Parameter analysis of algorithm 1:(a) the accuracies of 3 models under our attack with different step size; (b) the accuracies of 3 models under our attack with different step number.}
    \label{fig3}
\end{figure*}

\textbf{Step size}
We vary the step size from 0.01 to 0.1 in a granularity of 0.01. The number of step is set to 500 to ensure the convergence of the attack. The accuracy of three models are illustrated in Figure \ref{fig3}a. As can be observed, the robustness of the label smoothing model is significantly affected by the step size in comparison to two adversarial training models. Larger step size brings better attack effect.

\textbf{Step number}
We vary the number of steps from 50 to 500 in a granularity of 50 to test . Here we fix the step size to 0.01 and 0.1. As can be observed in Figure \ref{fig3}(b), Under the same setting, the convergence speed of the two adversarial training models is faster, while the accuracy of label smoothing model has been declining with the increase of step number.

In general, Under the same settings, the step size and the step number are the key to the success rate of model attack, which also supports the previous conclusion.

\subsection{Transferability}
A significant requirement for robustness is that it is not vulnerable to transfered attack. We test the transferability of the label smoothing model. Attack method is considered to have good transferability When the adversarial example generated on the original model can also mislead the target model. Using a natural training model as the source model and a 0.9 smoothing factor model as the target model, we test the 
robustness of label smoothing model under transfered attack. For comparison, two adversarial training models are tested for the target model. The results are shown in the Table \ref{tab6}.

\begin{table}[ht]
    \caption{The transferable accuracy of 3 models with different attack methods. Source model is a naturally-trained model.}
    \label{tab6}
    \centering
    \begin{tabular}{c|ccc}
    \toprule
         Attack Methods& Madry& Zhang&LabelSmoothing \\
         \midrule
         FGSM& 84.75 & 86.11&50.31\\
         PGD& 85.15 & 88.63&8.15\\
         CW& 85.19 & 87.90&9.23\\
         \bottomrule
    \end{tabular}
\end{table}

The results shown in the Table \ref{tab6} manifest that the label smoothing model is vulnerable to attacks from naturally trained model while two adversarial training model while the other two models are completely uninjured. This means that the change of one-hot label to label smoothing model alters the direction of gradient, making it incorrect to generate adversarial example in most cases. However, 
adversarial examples can still be found by other strategies to adjust the direction. Conversely, adversarial training models ensure that most examples cannot find adversarials, no matter how you change the direction of the perturbation.

\section{Related Work}
Jiang et al.\cite{jiang2020imbalanced} in their recent paper identify "imbalanced gradients", a new situation where traditional attacks such as PGD can fail and produce overstimated adversarial robustness. They state that the phenomenon of imbalanced gradients occures when the gradient of one term of the margin loss dominates and pushes the attack towards to a suboptimal direction. They mention that label smoothing causes imbalanced gradients but they do not explain how label smoothing causes imbalanced gradients. We think using imbalanced gradient to explain the reason is quite absolute and our study reveals robustness of the label smoothing model in more detail.

\section{Discussion}
Taken all results together, We can empirically conclude that label smoothing does not bring real robustness or stable robustness. It somehow bypasses most gradient-based attacks or exploits the weaknesses in these attacks. Increasing the step size, increasing the restart step, using an initialization strategy away from the initial image, and so on can greatly reduce the robustness of label smoothing model and even completely overwhelm the model. In other words, label smoothing makes the gradient-based attack method converge to the local sub-optimal. 

However, These shortcomings do not mean the label smoothing model has no practical effect. First, label smoothing is an extremely fast training method with little additional training expense; second, it is the most accurate method on clean samples in all defense methods, bypassing most of the currently widely used attacks without reducing the original accuracy of the model. Therefore, label smoothing is worth trying in practical situations where a relatively low level of defense needs to be achieved quickly and easily.

On the other hand, research on label smoothing reveals many problems in the current research of adversarial example. First of all, most of the current attack methods are not suitable or applicable to assessing the robustness of models, such as PGD, because the target of optimization does not guarantee the generation of adversarial example, so that the sub-optimal of algorithm convergence is not adversarial. In addition, most papers use the CW attack method to assess robustness, but only use margin loss instead of the original algorithm. From our experimental results, it can be seen that there are completely two different results.

So far, our work has explored various properties of the label smoothing model, but we have not found out the reason why label smoothing brings these changes to the model. At the same time, there is no efficient counter attack against the label smoothing model. If we can find a quick attack method, it will mean that we can take advantage of his shortcomings to produce a more efficient attack algorithm, and apply a stronger attack algorithm to adversarial training to produce a more robust model. We will focus on exploring the essence of the effect of label smoothing in future work, and we also hope that the research community will be committed to finding efficient and effective attack methods to evaluate the robustness of the model.

\section{Conclusion}
In this paper, we studied the effect of label smoothing during training on the model's adversarial robustness. We theoretically analyze why label smoothing invalidate most gradient-based attacks and evaluate the robustness of the label smoothing model in various experimental settings. We conclude that the robustness produced by label smoothing is incomplete; increasing the step size, adding restart operation, using different initialization strategies and so on, is the key to breaking through the label smoothing model. In general, label smoothing does not bring real robustness or stable robustness. Our research reveals that it is an urgent problem to propose valid and more efficient attack methods in adversarial example research.

\bibliographystyle{unsrt}
\bibliography{ref}

\end{document}